\documentclass[manuscript]{acmart}

\usepackage{subfig}
\usepackage{booktabs}
\usepackage{fancyhdr,graphicx,amsmath}
\usepackage[ruled,vlined]{algorithm2e}
\usepackage[inline]{enumitem}
\usepackage{wrapfig}
\DeclareMathOperator*{\argmax}{arg\,max}

\AtBeginDocument{%
  \providecommand\BibTeX{{%
    \normalfont B\kern-0.5em{\scshape i\kern-0.25em b}\kern-0.8em\TeX}}}

\begin{document}

\title{Counterfactual Contextual Multi-Armed Bandit: a Real-World Application to Diagnose Apple Diseases}


\author{Gabriele Sottocornola}
\authornote{Corresponding author.}
\email{gsottocornola@unibz.it}
\affiliation{%
  \institution{Free University of Bozen-Bolzano}
  \city{Bolzano}
  \country{Italy}
}

\author{Fabio Stella}
\email{fabio.stella@unimib.it}
\affiliation{%
  \institution{University of Milano-Bicocca}
  \city{Milano}
  \country{Italy}
}

\author{Markus Zanker}
\email{markus.zanker@unibz.it}
\affiliation{%
  \institution{Free University of Bozen-Bolzano}
  \city{Bolzano}
  \country{Italy}
}

\renewcommand{\shortauthors}{Sottocornola et al.}
\renewcommand{\shorttitle}{Counterfactual CMAB Application on Apple Disease Diagnosis}

\begin{abstract}

Post-harvest diseases of apple are one of the major issues in the economical sector of apple production, causing severe economical losses to producers. Thus, we developed \textit{DSSApple}, a picture-based decision support system able to help users in the diagnosis of apple diseases. Specifically, this paper addresses the problem of sequentially optimizing for the best diagnosis, leveraging past interactions with the system and their contextual information (i.e. the evidence provided by the users).
The problem of learning an online model while optimizing for its outcome is commonly addressed in the literature through a stochastic active learning paradigm - i.e. \textit{Contextual Multi-Armed Bandit (CMAB)}. This methodology interactively updates the decision model considering the success of each past interaction with respect to the context provided in each round. However, this information is very often partial and inadequate to handle such complex decision making problems. On the other hand, human decisions implicitly include unobserved factors (referred in the literature as \textit{unobserved confounders}) that significantly contribute to the human's final decision. In this paper, we take advantage of the information embedded in the observed human decisions to marginalize confounding factors and improve the capability of the CMAB model to identify the correct diagnosis. Specifically, we propose a \textit{Counterfactual Contextual Multi-Armed Bandit}, a model based on the causal concept of counterfactual. The proposed model is validated with offline experiments based on data collected through a large user study on the application. The results prove that our model is able to outperform both traditional CMAB algorithms and observed user decisions, in real-world tasks of predicting the correct apple disease.
\end{abstract}




\maketitle

\section{Introduction}
\label{sec:intro}

The domesticated apple (\textit{Malus x domestica}) is the third most produced fruit in the world (behind bananas and watermelons) with an amount of more than 86 million metric tons in 2018 \cite{fruit-prod}. In 2019, annual worldwide sales of apples were valued at around 7 billion USD \cite{apples-export}. Apple trees are indeed the most common temperate fruit tree species, since their fruits can be stored for prolonged periods of time under controlled atmosphere conditions. However, physiological disorders and pathogenic microorganisms can deteriorate the quality and quantity of the production during storage, and lead to considerable economic losses \cite{sutton:2014}. For example, in Northern Europe, storage losses due to pathogenic microorganisms were estimated to reach up to 10\% in integrated production and up to 30\% in organic production \cite{Maxin2014}. 
Therefore, an effective Decision Support System (DSS), that we referred as \textit{DSSApple}, able to timely diagnose occurring diseases and disorders on harvested apples is crucially important for this agricultural sector. For instance, it depends on the exact pathogen species to decide on the right strategy for immediate damage containment and/or to recommend a plant protection scheme for the following year. In order to reliably determine the nature of the disease, several macroscopic symptoms, such as appearance, color, texture and consistency of the rot need to be considered. Hence, the system should provide a practical interface to elicit user knowledge on the symptoms she might notice on a diseased apple, and a reasoning system to adapt the diagnosis to the user feedback. Nevertheless, the capability of the system to distinguish different candidate diseases relies on subtle differences in the symptoms' appearance that a non-expert user (or even an expert one) may failed to identify. Thus, another important feature for DSSApple is to correct user (imprecise) feedback in order to increase the accuracy of the suggested diagnosis.

In this work, we therefore investigate a component of such a DSS, namely, a method to interactively predict the best diagnosis, by leveraging past user interactions with a sequence of infected apples. The proposed model is based on the framework of Contextual Multi-Armed Bandit (CMAB) and on counterfactual causal inference. The former enables to process the observed sequential interactions of users with the diagnostic tool, the latter allows to extrapolate the bias in human decision making (the so-called \textit{unobserved confounders}) and improve the online diagnostic model.
The main contribution of this paper is to empirically assess the performance of the causal approach in a CMAB scenario, for the task of apple disease classification. Specifically, we base our work on the theoretical results presented so far in the literature \cite{agrawal:2013, bareinboim:2015, lee:2018}, in order to implement a novel \textit{Counterfactual CMAB} algorithm, called \textit{Counterfactual Thompson Sampling (CF-TS)}, tailored to our problem. We demonstrate the effectiveness of the proposed algorithm by testing it on real-world data related to the diagnosis of post-harvest apple diseases. The offline evaluation was derived from a large user study, in which we collected user interactions with DSSApple. The participants of the study were asked to diagnose a set of diseased apples images, crafted by domain experts in the laboratory, by using DSSApple application. The simulation derived from this data, showed that CF-TS significantly outperforms other decision models, i.e. observational and CMAB models, in the task of sequentially diagnosing the correct apple disease.

This work represents a further step in the development of an effective and practical DSS that aim at helping expert and non-expert user in the diagnosis of post-harvest diseases of apple. In particular, the fundamental insight gained in this research is that an active learning classifier based on a counterfactual CMAB algorithm is able to outperform both the observed user decision model and a state-of-the-art CMAB policy. The performances of the improved diagnostic tool have been proved to reach an accuracy greater than $40\%$ in a real-world setting with $5$ possible diseases, just leveraging users' interactions with the application showing pictures of macroscopic symptoms. This result could be easily converted in a immediate reduction of the economical losses in apple production due to undetected post-harvest infections.

\section{Related Work}

Since early years, \textit{Multi-Armed Bandit (MAB)} \cite{sutton:1998} algorithms have been largely applied and proved to be effective in the task of optimizing sequential decisions. They achieve an optimal exploration-exploitation trade-off by minimizing the cumulative regret over a finite time horizon \cite{auer:2002}. Along with theoretical analysis of upper bound regret, some works tested the effectiveness of MAB on synthetic and real-world classification tasks \cite{vermorel:2005, kakade:2008, pandey:2007}. \textit{Thompson Sampling (TS)} policy for Beta-Bernoulli bandit, extended with regularized logistic regression, was evaluated in the applications of online advertising and news recommendation by Chapelle and Li \cite{chapelle:2011}. Moreover, \textit{Contextual Multi-Armed Bandit (CMAB)} was successfully applied throughout the years. This methodology provides a more realistic and adaptive setup in which each instance is processed with its side information, referred to as context. The learning process for identifying the best arm is a function of the context at each iteration \cite{langford:2007}.
A well-known example of a real-world application of CMAB is represented by Li et al. \cite{li:2010}. The authors proposed an improved version of the \textit{Upper Confidence Bound (UCB)} algorithm with disjoint and hybrid linear models. Furthermore, an effective offline evaluation of the bandit policies was performed on the click-through-rate of the news shown on the \textit{Yahoo!} homepage. Agrawal and Goyal \cite{agrawal:2013} illustrated a compact implementation of TS for the Gaussian bandit with contextual information, together with a proof of bounded regret. These works systematically evaluate the CMAB algorithms proposed so far and an additional novel one for document classification \cite{agarwal:2014}.

Recently, the so-called ``causal revolution'' \cite{pearl:2000} invested the field of reinforcement learning and multi armed bandit. Initially, the complex causal relationships in the application scenarios of online computational advertising were investigated by Bottou et al. \cite{bottou:2013}. In their seminal work on causal MAB, Bareinboim et al. \cite{bareinboim:2015} presented a causal approach to the Beta-Bernoulli TS algorithm. In particular, they illustrate a novel algorithm able to de-bias the bandit learning process from unobserved confounders, influencing both the arm selection and the observed outcome (i.e. the reward). The method proved to outperform the non-causal counterparts in terms of measured regret on synthetic examples. A more extensive description of the theory behind the MAB with confounders factors is illustrated in \cite{forney:2017} and in \cite{zhang:2017}. This work exploits transfer learning in order to combine observational, experimental, and counterfactual data through structural causal models. In their extended work, Lee and Bareinboim \cite{lee:2018} enhance the concept of causal multi-armed bandit by fusing them with structural causal graphs. MAB outcome is influenced by both, unobserved confounders and interventional action by the agent. The POMIS algorithm (partial order of minimal intervention set) has been proposed to find the best arm solution. Again, experiments are conducted based on a simulation of pre-defined tasks.
More recently, an application of non-contextual causal MAB on simulated e-mail data was presented by Lu et al. \cite{lu:2020}. The authors of \cite{dimakopoulou:2019}, instead, proposed a CMAB algorithm balanced with causal inference and they tested it on a benchmark of 300 multi-class classification datasets.

\section{Application Design}
\label{sec:design}

\begin{figure}[htb]%
    \centering
    \subfloat[\centering]{{\includegraphics[width=7cm]{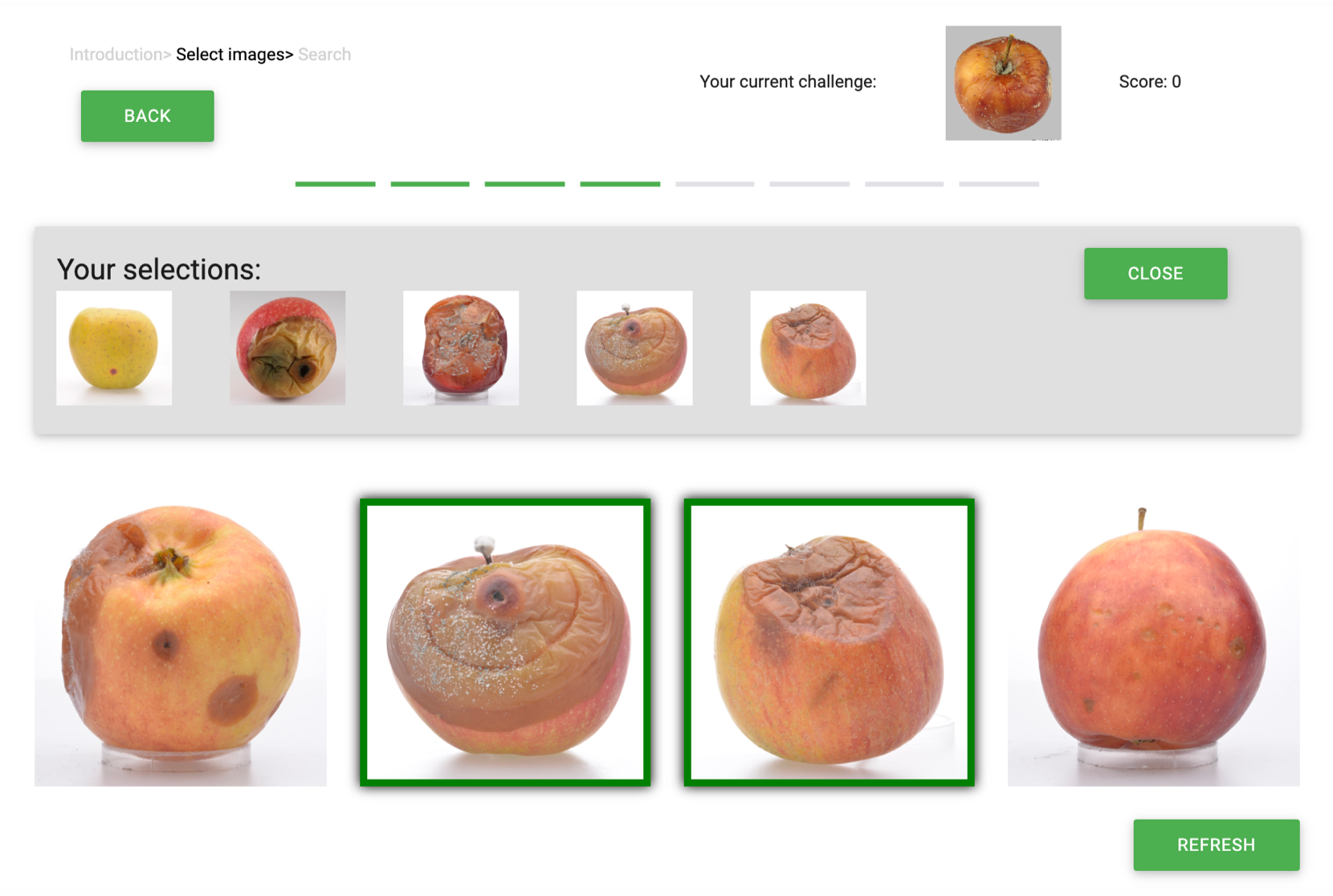} \label{fig:dss-interface}}}%
    \qquad
    \subfloat[\centering]{{\includegraphics[width=7cm]{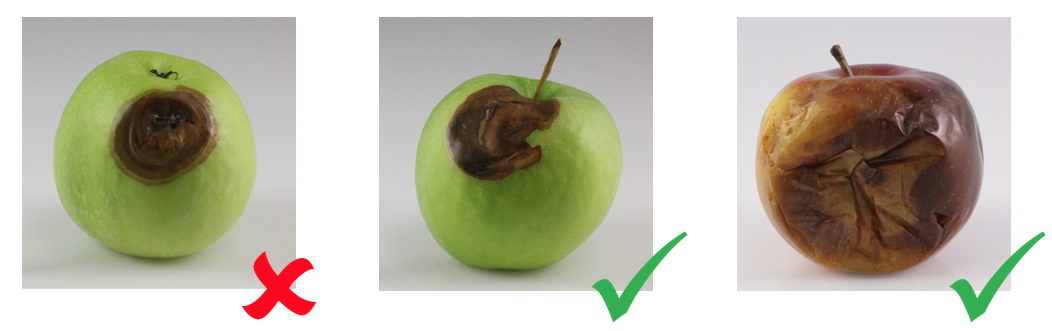} \label{fig:apple-comp}}}%
    \caption{(a) The DSSApple interface. (b) An example of how difficult could be to identify the correct disease by its macroscopic symptoms. The left-most apple is infected by Neofabraea, while the others are infected by Alternaria.}%
    \label{fig:design-choice}%
\end{figure}

For the task of diagnosing post-harvest disease of apple from its macroscopic symptoms, the employment of fully-automated machine learning techniques for image recognition appears to be insufficient up to now. The intra-disease variance is very high: the same pathogen induces different symptoms on different species, also based on the progression of the diseases (i.e. days after an infection). At the same time for a non-expert evaluation - and even for experts without a microscopic or microbiological analysis - it is really difficult to understand the subtle differences of symptom appearances just by observing images of external symptoms, particularly at early stages of infection.
In Figure \ref{fig:apple-comp} we show three instances of external symptoms, that clearly highlight the difficulty of the classification task. The two symptoms that look most similar - given also that they appear on the same apple cultivar - are in fact manifestations of two different diseases (\textit{Neofabraea} and \textit{Alternaria}). On the other hand, two examples of \textit{Alternaria} symptoms appear to be largely different, since they manifest themselves on different cultivars and at different stages of the infection.

In order to tackle this challenging problem, we developed a decision support system, named \textit{DSSApple}, designed to be an easy-to-use web application that helps expert (e.g. storage workers, researchers) and non-expert users to perform in-field diagnosis of apple diseases from macroscopic symptoms (i.e. without any microscopic investigation). Given the discussed peculiarity of the domain, we choose to design the application as a human-in-the-loop interactive DSS. The design of this prototype was mainly inspired by the work by Pertot et al. \cite{pertot2012}. The interaction of the user with the system is conducted simply by clicking on pictures (referred to as \textit{feedback images}), representing the symptoms' variety of different diseases at different stages of infection. We conceptualized an interactive session with the system as a sequence of rounds. At each round of interaction the user provides immediate feedback on a small set of images, depicting disease symptoms, based on the perceived similarity with an actually diseased target apple. We proved that this modality increases the system usability and alleviates the cognitive overload of the user \cite{nocker:2018}. Feedback images have been produced by taking pictures of sampled apples from different storage houses, where the ground truth (i.e. the actual disease) has been determined by domain experts in laboratory using gene sequencing of spores. At the current stage, DSSApple includes 5 different pathogens, namely, \textit{Alternaria}, \textit{Botrytis}, \textit{Mucor}, \textit{Neofabraea}, and \textit{Penicillium}, represented by a pool of 30 feedback images each. The limited total number of labelled high-quality images is another crucial factor that discourage the usage of machine learning techniques for image classification.
Figure \ref{fig:dss-interface} depicts a round of interaction, where users can provide feedback on any number of small-scale feedback images, before submitting their choices to the system. At the end of each round, the system automatically reloads alternative images based on the feedback provided by users. Different reloading policies have been tested to better adapt the new feedback images to previous user feedback and, hence, increase the diagnostic accuracy of the system \cite{sottocornola:2020}. After a fixed number of rounds, DSSApple stops feedback collection and suggests a set of candidate diseases that are ranked based on the number of coherent user feedback on the symptom images belonging to each disease. Finally, the user can communicate her final choice to the system, by selecting one among the suggested diseases as the final diagnosis for the target infected apple.

In this paper, we focus on the component of DSSApple responsible for the diagnosis - i.e. the classification of user feedback into one or more suggested diseases. In particular, we treat each user session with the system as an event, and we aim at sequentially optimizing each event for the best diagnosis. For each event, we exploit all the user feedback collected by the system in that session, as well as the past user interactions, in order to provide more reliable diagnosis and, ultimately, improve the accuracy of DSSApple. Thus, to achieve this goal we propose a novel counterfactual contextual multi-armed bandit algorithm, adapted to the problem of diagnosing post-harvest disease of apple.

% \begin{wrapfigure}{r}{0.5\textwidth}
%   \begin{center}
%     \includegraphics[width=0.4\textwidth]{img/dss-interface.png}
%   \end{center}
%   \caption{Bad Apple Challenge application interface.}
%   \label{fig:pic_search}
% \end{wrapfigure}

\section{Methodology}
\label{sec:meth}

We combine disjoint linear CMAB for classification \cite{li:2010}, with specific reference to the Thompson Sampling (TS) policy \cite{agrawal:2013}, to the powerful concept of counterfactual \cite{pearl:2000}. In particular, counterfactual modeling and reasoning allow us to effectively leverage on human decisions (or intuitions) when a given arm has to be selected.

\begin{figure}[htb]%
    \centering
    \subfloat[\centering]{{\includegraphics[width=5cm]{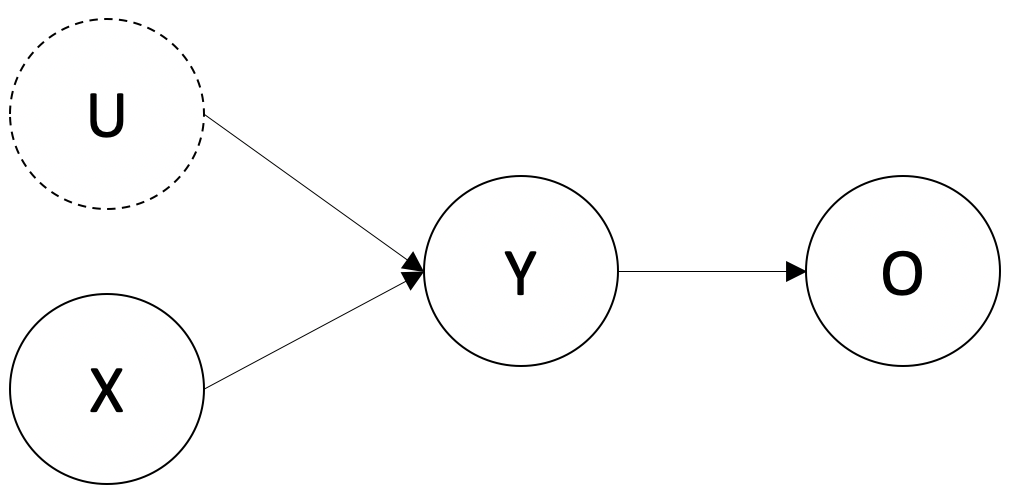} \label{fig:causal-model1}}}%
    \qquad
    \subfloat[\centering]{{\includegraphics[width=6cm]{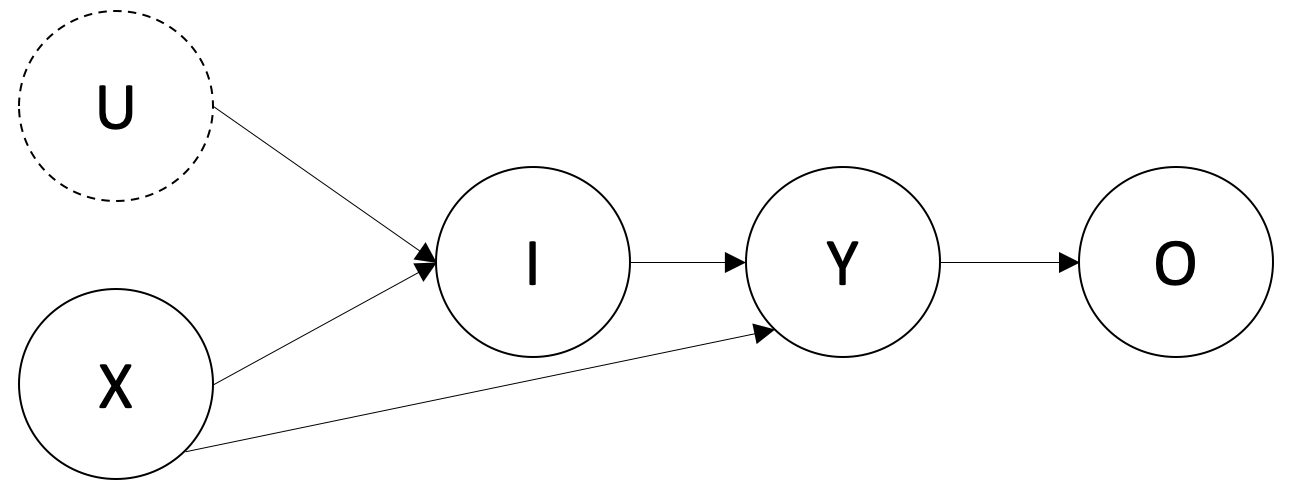} \label{fig:causal-model2}}}%
    \caption{Causal model for the standard decision making scenario (a) and the counterfactual decision making (b).}%
    \label{fig:causal-models}%
\end{figure}%
Consider the causal diagram in Figure \ref{fig:causal-model1} depicting the decision making process in a MAB scenario, where node \textit{X} represents the contextual information, i.e. the information explicitly available to the decision maker. In contrast, node \textit{U} represents the unobserved confounders, i.e. the hidden information which influences the final decision. Node \textit{Y} represents the final decision or arm selection (e.g. the diagnosis), while node \textit{O} represents the outcome of that decision (e.g. whether the diagnosis \textit{Y} was correct or not). We are aware of the fact that in a more general scenario the decision outcome \textit{O} is directly influenced by contextual and unobserved factors as well. We decided to stick with this simplified causal model given that it better mimics our application scenario (i.e. in a diagnostic scenario, the correctness of the diagnosis \textit{O} is conditionally independent on both the observed variable \textit{X} and unobserved variable \textit{U} given the diagnosis \textit{Y}). A CMAB model trained on this decision problem learns only from the observed contextual information \textit{X}. This is a strong limitation because it excludes relevant information that would be necessary in order to make an effective decision. On the other hand, humans make decisions based on a mixed interpretation of the observed context and the unobserved confounders (e.g. the diagnosis of a medical doctor is based on both structured knowledge on the observed evidence and the intuition over past experience) \cite{forney:2017}. Nevertheless, confounders could hardly be directly incorporated into the decision model, due to their nature. Therefore, we propose to exploit the human decision as a proxy to implicitly access the unavailable information included in the unobserved confounders. In this situation, the decision making scenario is represented by the causal graph in Figure \ref{fig:causal-model2}. The intuition of the user (observed node \textit{I}) is conditionally dependent on both contextual information and confounders factors. The final decision (arm selection) is thus conditioned on node \textit{I} and \textit{X}, including all the available information in the decision process.

\begin{algorithm}[]
\SetAlgoLined
 Input: $B_{i,j} = I_d, \hat{\mu}_{i,j} = 0_d, f_{i,j} = 0_d$\;
 \ForEach{t = 1, 2, \ldots T}{
  Receive a context $x(t)_d$ and an arm intuition $i(t)$\;
  \ForEach{y = 1, 2, \ldots A}{
  Sample $\Tilde{\mu}_{i(t),y}(t)$ from $N(\hat{\mu}_{i(t),y}, B_{i(t),y}^{-1})$\;
  Compute $E[r_{i(t),y}(t)] = x(t) \Tilde{\mu}_{i(t),y}(t)$\;
  }
  Play arm $a(t) := \argmax_y E[r_{i(t),y}(t)]$ and observe reward $r(t)$\;
  $B_{i(t),a(t)} = B_{i(t),a(t)} + x(t) x(t)$\;
  $f_{i(t),a(t)} = f_{i(t),a(t)} + x(t) r(t)$\;
  $\hat{\mu}_{i(t),a(t)} = B_{i(t),a(t)}^{-1} f_{i(t),a(t)}$\;
 }
 \caption{Counterfactual Thompson Sampling (CF-TS) for Contextual Bandit Diagnosis}
 \label{alg:cf-ts}
\end{algorithm}

The complete procedure of the \textit{Counterfactual Thompson Sampling (CF-TS)} is illustrated in Algorithm \ref{alg:cf-ts}. The structure of the algorithm is similar to the one of a Gaussian CMAB with linear payoff and $A$ arms, where $A$ corresponds to the number of possible actions (or interventions). In this case, following the Thompson Sampling policy, the exploration-exploitation trade-off is provided by sampling from a multi-variate Gaussian $N(\hat{\mu}, B^{-1})$ with $d$ components. The parameters update (last 3 lines) is obtained through the Bayesian update described in \cite{agrawal:2013}. The novelty of the proposed algorithm stays in its capability of computing the expected reward (and thus the final decision) as a function of both the contextual information and the decision intuitively made by the user. This concept follows the causal relationships depicted in Figure \ref{fig:causal-model2}, where $Y = f(X,I)$. In more details, at time $t$, when a decision needs to be taken, the algorithm receives the context vector $x(t)$, as well as the arm intuition provided by the human decision maker $i(t)$. Following the notation in \cite{bareinboim:2015}, the reward computation $E[r_{i(t),y}(t)] = x(t) \Tilde{\mu}_{i(t),y}(t)$ is equivalent to compute $E[r(t)_y|I=i(t), X=x(t)]$, the expected reward at time $t$ for selecting arm $y$ given the contextual information $X = x(t)$ and the intuition of the user towards arm $I = i(t)$. In other words, this formula answers the counterfactual question: ``What have been the expected reward had I pull arm $y$ given that I am about to pull arm $i(t)$?''. Finally, the algorithm finds the best arm $a(t) = y$ which maximizes the expected reward $E[r(t)]$ for each decision/arm $y$.

\section{Experimental Setup}
\label{sec:exp}

% The offline experiments are performed on the data derived from the \textit{Bad Apple Challenge} \cite{sottocornola:2020}. This was a gamified user study conducted in order to validate the effectiveness of an interactive application for guiding the user through the diagnosis of an apple disease. Users were shown pictures of diseased apples, that served as the target that needs to be diagnosed. During the challenge users navigated the application by selecting images (referred as \textit{challenge images}) of other diseased apples, showing symptoms similar to the one she received as a target (Figure \ref{fig:pic_search}). This allowed the system to elicit user feedback and provide a ranked list of candidate diseases. Finally, users had to take the final decision and select the disease they considered the correct one for the target apple.
We designed a gamified user study, that we called \textit{Bad Apple Challenge}, in order to validate the effectiveness of DSSApple \cite{sottocornola:2020}. The user, at each session, receives a target infected apple, for which she needs to diagnose the unknown disease. The participant navigates the DSSApple application (as described in Section \ref{sec:design}) in order to receive a ranked list of suggested diagnosis. At the end of the session, the user has to select the diagnosis she consider the correct one for the target apple. The user accumulates score points if the diagnosis is correct (i.e. the selected disease corresponds to the actual disease of the target apple). We administrated the Bad Apple Challenge in a controlled environment with $163$ non-expert participants (i.e. two cohorts of BSc Computer Science students).\\
We exploit the data collected through the Bad Apple Challenge to evaluate the effectiveness of the CF-TS algorithm, applied to the task of sequentially diagnosing apple diseases. Each challenge session is treated as a classification instance, where the feedback images selected by the user represent the context of the diagnosis. The final selection by the user is used as the intuition towards the diagnosis. The ground truth disease contributes to the computation of the reward. Thus, the sequential disease classification is converted into a multi-armed bandit problem where each candidate disease is mapped to an independent arm. Given that the number of candidate diseases is $5$, we deal with a $5$-armed bandit problem. A reward of $1$ is assigned if the selected arm corresponds to the true disease of the target apple, $0$ otherwise.
Given the set of feedback images clicked during a challenge we extract two types of context:
\begin{enumerate*}
    \item \textbf{Image-based Context (ImgCtx)}: The context of the diagnosis is computed as the sum of the pixel-wise representation of every selected feedback image. In order to compress the feature space and reduce the noise we apply to each unfold image vector a PCA with $64$ dimensions.
    \item \textbf{Similarity-based Context (SimCtx)}: The context of the diagnosis is computed as the sum of the similarity vectors of each selected feedback image. The similarity vector is derived from user interactions as the relative frequency of an image being co-clicked with others during the same challenge. Again the feature space is reduced to the $64$ principal components through PCA.
\end{enumerate*}

The total number of challenge sessions performed in the user study is $515$, which is too few to evaluate the convergence of bandit algorithms. We apply a fair and effective technique to simulate more instances. We replicate each session $r$ times, but, in each replica, every selected feedback image has a dropout probability $d > 0$ to be discarded. Thus, the context computation has some degree of randomness and we ensure each replicated instance to be different from the others. After the dataset is constructed we apply a random shuffling. Furthermore, to obtain more robust results, we replicate this process $n$ times, each time generating a different enhanced dataset to be used in the comparative offline evaluation. In our experiments, we set  $r = 2, 4, 6$ to test each algorithm on a time horizon of $t = 1000, 2000, 3000$ observations, respectively. The dropout probability is always set to $d = 0.2$ and the number of repetitions of the whole process is set to $n = 100$.
In order to validate the performance of the proposed algorithm we compare it along with 3 baseline selection models. The methods tested in the experiments are:
\begin{enumerate}
    \item \textbf{Counterfactual Thompson Sampling (CF-TS)}: The proposed method described in Section \ref{sec:meth}.
    \item \textbf{Observational (Obs)}: The observed decision model, i.e. the actual selection of each user at the end of the session.
    \item \textbf{Thompson Sampling (TS)}: The standard implementation of a set of disjoint Thompson Sampling classifiers with linear payoff, as described in \cite{li:2010, agrawal:2013}.
    \item \textbf{Extended Thompson Sampling (ExtTS)}: The standard TS algorithm with an extended context. Namely, the user intuition is encoded in a one-hot vector and appended at the context vector. This baseline guarantees that the TS algorithm processes the same amount of information as the CF-TS method.
\end{enumerate}

\section{Results}
\label{sec:res}

In the following we summarize the results obtained in the offline evaluation\footnote{The full code of the experimental evaluation of the implemented methodology is available at \url{https://github.com/endlessinertia/causal-contextual-bandits}}. The results reported in this section are all derived as an average over the $100$ replications of each experiment.
\begin{figure}[!htb]%
    \centering
    \subfloat[\centering]{{\includegraphics[width=7cm]{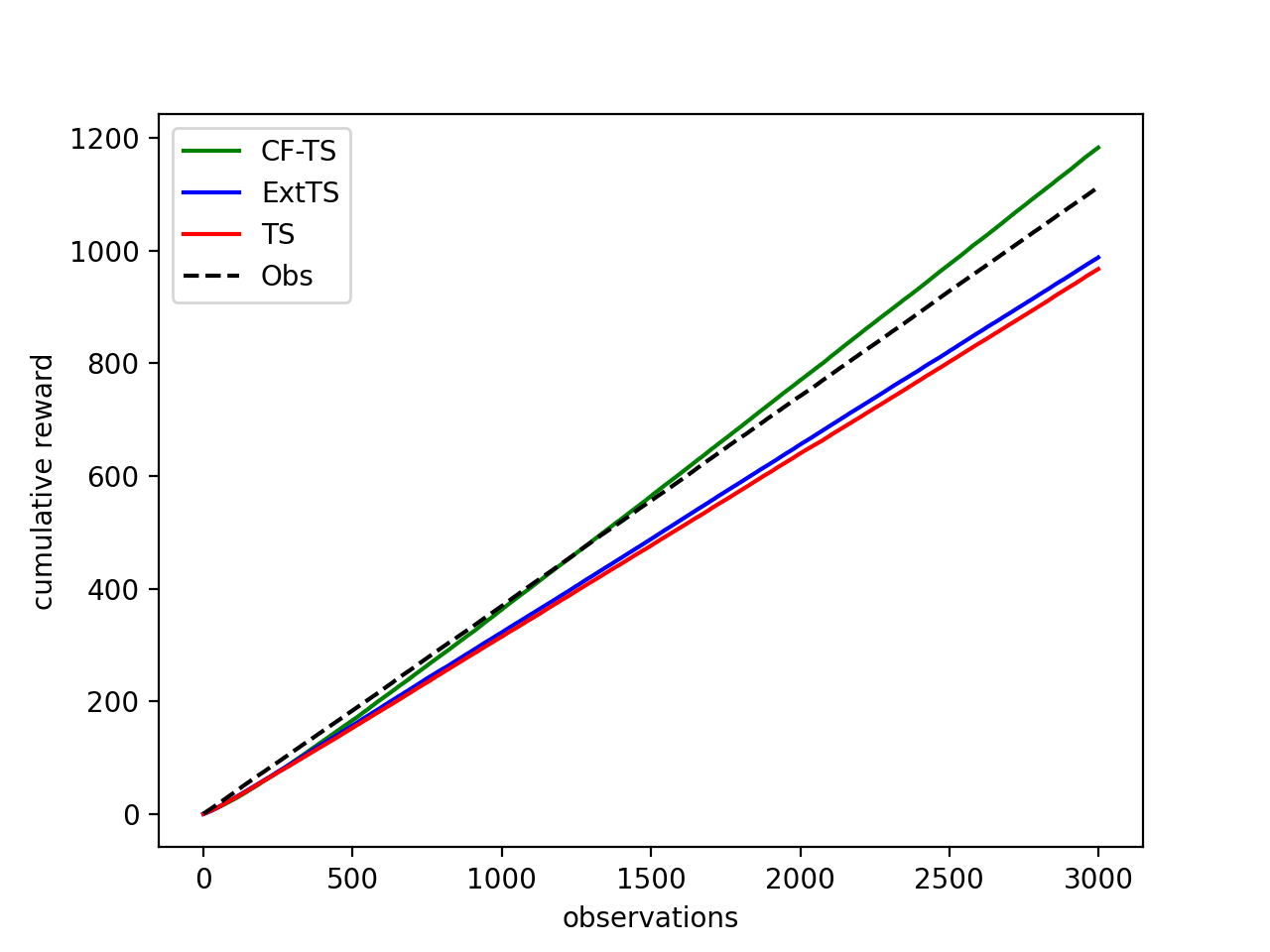} \label{fig:3000_rew_img}}}%
    \qquad
    \subfloat[\centering]{{\includegraphics[width=7cm]{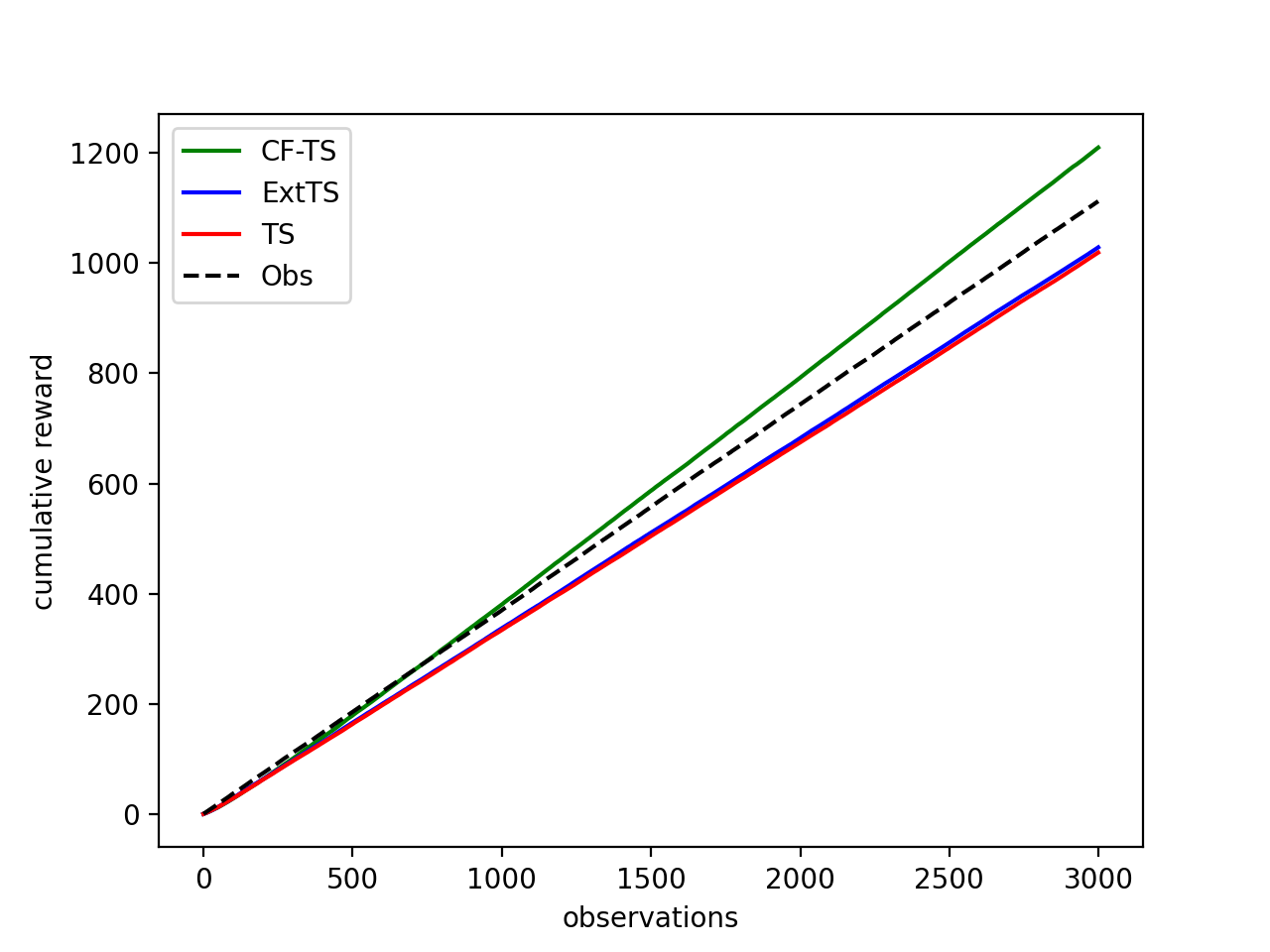} \label{fig:3000_rew_sim}}}%
    \caption{Cumulative reward after 3000 interactions, considering the image-based context, \textit{ImgCtx} (a) and the similarity-based context, \textit{SimCtx} (b) for CMAB.}%
    \label{fig:3000rew}%
\end{figure}
The graphs in Figure \ref{fig:3000rew} depict the cumulative reward of every method with respect to the two context types, namely \textit{ImgCtx} (a) and \textit{SimCtx} (b). The time horizon represented in the graph is the largest one of $3000$ observations. The behavior of the methods after a smaller number of iterations (i.e. $t = 1000, 2000$) can be derived directly from these graphs.
In both situations, after a certain number of iterations, the \textit{CF-TS} method emerges as a clear winner, achieving the maximum reward among the competitors. Nevertheless, the algorithm needs a warm-up period that depends on the provided context. In the case of \textit{ImgCtx}, \textit{CF-TS} needs around $1200$ iterations to achieve results comparable to the ones obtained by human selection. In the other case, with a more meaningful context derived by collaborative image similarity, the algorithm is faster in outperforming the observational model, around the $1000$-th iteration. Important to notice how the two TS baselines are never able to get closer to the performance achieved by the observational model, irrespectively on the number of observations they can process. 
%After an initial warm-up period that last around $500$ iterations in which the three CMAB models explore the different arms and achieve similar performances, their results start to diverge. The counterfactual model benefits from the new observed instances and increases its predictive capability, while TS and ExtTS stay constantly below the observational model. 
Another interesting aspect is that the extended model, which exploits a larger contextual information, registers a small gain in the case in which the original context is sub-optimal (namely for \textit{ImgCtx}); thus, an improved representation of the context can effectively boost the TS model. Nevertheless, the \textit{ExtTS} model remains mostly in line with standard \textit{TS} and, despite the augmented observational information, it is not able to capitalize on it and to achieve a performance comparable to the proposed \textit{CF-TS} model.

In Table \ref{tab:res-acc} we summarize the accuracy results of each algorithm in diagnosing the correct disease as a function of the number of observations (i.e. $t = 1000, 2000, 3000$) and the type of context provided as input (i.e. \textit{ImgCtx} and \textit{SimCtx}). The accuracy for an algorithm $m$ at time horizon $t$ is simply computed as $a^m(t) = 1/t \sum_{i=1}^t r^m(i)$, where $r^m(i)$ is the reward obtained by the algorithm $m$ at the $i$-th iteration. We also include the performance of a \textit{ZeroR} baseline, as a reference point. The \textit{ZeroR} classifier always assigns as diagnosis the most frequent true disease in the set.

\begin{table}[H]%
\centering
\begin{tabular}{@{}|c|cc|cc|cc|@{}}
\toprule
 & \multicolumn{2}{c|}{\textit{t = 1000}} & \multicolumn{2}{c|}{\textit{t = 2000}} & \multicolumn{2}{c|}{\textit{t = 3000}} \\
 & \textit{ImgCtx} & \textit{SimCtx} & \textit{ImgCtx} & \textit{SimCtx} & \textit{ImgCtx} & \textit{SimCtx} \\ \midrule
\textit{CF-TS} & 0.354* & 0.367* & \textbf{0.379*} & \textbf{0.392*} & \textbf{0.394*} & \textbf{0.403*} \\
\textit{ExtTS} & 0.322 & 0.345 & 0.330* & 0.341 & 0.330* & 0.343 \\
\textit{TS} & 0.317 & 0.338 & 0.323 & 0.340 & 0.322 & 0.340 \\
\textit{Obs} & \textbf{0.371*} & \textbf{0.371} & 0.371* & 0.371* & 0.371* & 0.371* \\ \midrule
\textit{ZeroR} & 0.221 & 0.221 & 0.221 & 0.221 & 0.221 & 0.221 \\ \bottomrule
\end{tabular}%
\caption{Accuracy at different time horizons $t = 1000, 2000, 3000$, considering \textit{ImgCtx} and \textit{SimCtx} for CMAB and baseline selection models. * indicates improvements on other methods in the same column being significant at p-value < 0.01 on paired samples t-test.}
\label{tab:res-acc}%
\end{table}

The results of Table \ref{tab:res-acc} clearly highlight the different behaviours of the 4 selection models. The observational model outperforms its counterparts in the first two scenarios in which the total number of iterations is set to $t = 1000$. A general trend for the TS-like algorithms is to benefit from the more informative context, showing an increment between $0.8\%$ and $2.3\%$ of accuracy, with respect to the same time horizon. In the experiments with $t = 2000, 3000$ and \textit{ImgCtx}, \textit{ExtTS} significantly improves the accuracy w.r.t. \textit{TS} from $32\%$ to $33\%$, by exploiting the enhanced context. The same does not hold true for the \textit{SimCtx}, where the gain is not significant. This is again due to the fact that the \textit{SimCtx} provides a more reliable information than the \textit{ImgCtx}, which gets more benefit from the context enrichment. Finally, it is interesting to notice how, after $2000$ iterations the two TS baselines reach their upper bound in the predictive accuracy (i.e. no improvement is registered in the case of $t = 3000$). Nevertheless, the proposed \textit{CF-TS} is capable to constantly improve its performance at the different cutoffs, reaching a maximum of $40\%$ accuracy after $3000$ iterations for the \textit{SimCtx} context ($+3\%$ on the observational decision model).

% \begin{figure}[htb!]%
%     \centering
%     \subfloat[\centering]{{\includegraphics[width=6cm]{img/3000_arms_imgs.png} \label{fig:3000_arms_img}}}%
%     \qquad
%     \subfloat[\centering]{{\includegraphics[width=6cm]{img/3000_arms_int.png} \label{fig:3000_arms_sim}}}%
%     \caption{Arms selection distribution after 3000 interactions, considering the image-based context (a) and the similarity-based context (b) for CMAB.}%
%     \label{fig:3000arms}%
% \end{figure}
% In Figure \ref{fig:3000arms} we represent the distribution of the arm selection compared to the true and the observed one, at time horizon $t = 3000$. Important to notice that the arms are ordered on the x-axis by the frequency of selection in each experiment for each algorithm. Thus, the x-axis does not represent the id of the arm but just its ranking.
% These plots induce some consideration on the nature of the compared algorithms. Despite the fact that the true class distribution is almost perfectly balanced, both the CMAB baselines select around $90\%$ of the times one of the two most frequent arms. This is mainly due to the linearity in the payoff computation, which does not allow a learning mechanism complex enough to avoid the ``rich-gets-richer'' effect, after the first rounds of exploration.
% On the other hand, the observed user selections over the diseases are distributed similarly to the true ones. This allows the CF-TS to perform a more heterogeneous arm selection which is closer to the true arm distribution.

\section{Conclusions}
\label{sec:conc}

We introduced an interactive decision support system, called \textit{DSSApple}, for tackling the real-world problem of diagnosing post-harvest diseases of apple.
Specifically, in this paper we presented a novel Counterfactual CMAB algorithm, that is responsible to improve the diagnostic performance of the model, by sequentially leveraging users' interactions with the system. The algorithm, called \textit{Counterfactual Thompson Sampling (CF-TS)}, exploits human decision (i.e. the intuition) and the contextual information (i.e. the evidence) to compute the optimal counterfactual choice. We evaluated the effectiveness of CF-TS by comparing it with the performance of the observed human decisions and two natural Thompson Sampling baselines. In the offline experiments, simulated from data collected in a large gamified user study, CF-TS was able to outperform all the baselines and achieve significant improvements in diagnosing the correct diseases. In future work, the application will be tested in a real-world environment by storage workers and quality control managers, in order to assess its diagnostic capability in-field. A crucial future extension of the presented application consists in incorporating a knowledge base that models domain expertise, to include richer information (e.g. inner/outer symptoms description, cultivar type, handling conditions, etc.) to increase the overall success rate and to be capable to better explain the final diagnosis and recommend suitable countermeasures. 
%Furthermore, we are going to test the presented counterfactual CMAB methodology in other real-world domain in which unobserved confounders play an important role in the decision making process (e.g. product recommendation). 

\bibliographystyle{ACM-Reference-Format}
\bibliography{DSSApple_CF-CMAB}

\appendix

\end{document}